%% file: main_ieee_format.tex
\input{glyphtounicode}
\documentclass[letterpaper, 10 pt, conference]{ieeeconf}  

\IEEEoverridecommandlockouts                              

\usepackage[T1]{fontenc}
\usepackage[utf8]{inputenc}
\usepackage[natbib=true]{biblatex}

\input{config}
\input{macros}

\usepackage{amssymb}
\usepackage{graphicx}
\usepackage{cprotect}
\usepackage{xcolor}
\usepackage[
    bookmarks=false,
    pdfencoding=auto, 
    psdextra,        
    colorlinks=true, 
    linkcolor=blue, 
    urlcolor=blue, 
    citecolor=blue
]{hyperref}
\usepackage{cleveref}
\usepackage[noend]{algpseudocode}
\usepackage{algorithm,algorithmicx}
\usepackage{booktabs}
\usepackage{multirow}

\newcommand*\Let[2]{\State #1 $\gets$ #2}
 \algrenewcommand\algorithmicrequire{\textbf{Input:}}

\title{\LARGE \bf
Task Robustness via Re-Labelling Vision-Action Robot Data
}

\author{
  \textsuperscript{$\dagger$}Artur Kuramshin\textsuperscript{1,2}, \; Özgür Aslan\textsuperscript{1,2}, \; Cyrus Neary\textsuperscript{1,2,3}, \; Glen Berseth\textsuperscript{1,2}\\
  \thanks{\textsuperscript{1}Mila --- Quebec AI Institute, \textsuperscript{2}Université de Montréal, \textsuperscript{3}The University of British Columbia, \textsuperscript{$\dagger$}For correspondence and questions: \texttt{artur.kuramshin@mila.quebec}} 
}

\begin{document}
\maketitle
\thispagestyle{empty}
\pagestyle{empty}

\input{icra_submission/00_abstract}


\input{icra_submission/01_introduction}
\input{icra_submission/02_related_work}
\input{icra_submission/03_method}
\input{icra_submission/04_results}

\input{icra_submission/05_conclusions}

\input{icra_submission/06_appendix}
\vspace{-0.5cm}
\section*{ACKNOWLEDGMENTS}
We acknowledge the use of Claude Sonnet 4 (Anthropic) for assistance with editing and refining the written presentation of this work. All scientific content, methodology, and conclusions remain the original contributions of the authors.

We want to acknowledge funding support from Natural Sciences and Engineering Research Council of Canada, Fonds de recherche du Québec and The Canadian Institute for Advanced Research (CIFAR) and compute support from Digital Research Alliance of Canada, Mila IDT, and NVidia. We would also like to thank John Edwards Leadership Fund (CFI) for funding the purchase of hardware for this project.

\printbibliography





\end{document}

%% file: config.tex
\usepackage{xspace}

%% file: macros.tex
\newcommand{\methodName}{TREAD\xspace}

%% file: icra_submission/00_abstract.tex
\begin{abstract}

The recent trend in scaling models for robot learning has resulted in impressive policies that can perform various manipulation tasks and generalize to novel scenarios. However, these policies continue to struggle with following instructions, likely due to the limited linguistic and action sequence diversity in existing robotics datasets. This paper introduces \textbf{T}ask \textbf{R}obustness via R\textbf{E}-Labelling Vision-\textbf{A}ction Robot \textbf{D}ata (TREAD), a scalable framework that leverages large Vision-Language Models (VLMs) to augment existing robotics datasets without additional data collection, harnessing the transferable knowledge embedded in these models. Our approach leverages a pretrained VLM through three stages: generating semantic sub-tasks from original instruction labels and initial scenes, segmenting demonstration videos conditioned on these sub-tasks, and producing diverse instructions that incorporate object properties, effectively decomposing longer demonstrations into grounded language-action pairs. We further enhance robustness by augmenting the data with linguistically diverse versions of the text goals. Evaluations on LIBERO demonstrate that policies trained on our augmented datasets exhibit improved performance on novel, unseen tasks and goals. Our results show that TREAD enhances both planning generalization through trajectory decomposition and language-conditioned policy generalization through increased linguistic diversity. Project website: \href{https://akuramshin.github.io/tread}{https://akuramshin.github.io/tread}.

\end{abstract}

%% file: icra_submission/01_introduction.tex
\section{Introduction}

Recent trends in robot learning research have demonstrated the importance of data scale and diversity for learning more robust and capable robotic manipulation policies \citep{bharadhwaj2024roboagent, kim2024openvla, black2410pi0}. While the robotics community has made progress in expanding dataset sizes through collaborative efforts, these datasets continue to exhibit limited diversity along certain text and trajectory modalities \citep{o2024open, khazatsky2024droid, walke2023bridgedata}. This limitation manifests as a weakness of the current models to reliably follow text-based instructions \citep{pertsch2025fast, kim2025fine, liu2024rdt}.


The dataset modality imbalance persists due to the challenges of collecting diverse, real-world expert demonstrations at scale, often requiring multiple robots and months of operation to obtain even modest datasets \citep{brohan2022rt, khazatsky2024droid, fang2024rh20t}. However, these vision-based manipulation demonstrations have grown in length and complexity, covering many potential sub-goals per trajectory. 
While it is possible to manually review the thousands of demonstration videos and solicit human annotators, this will not scale with the demands of modern robotic learning systems. This raises the question: How can we systematically augment existing robotics datasets with greater language-action diversity in a scalable way?


Recent advances in large-scale vision-language models (VLMs) \cite{achiam2023gpt, bai2025qwen25vltechnicalreport, gemini25} offer a promising new approach to address the language-action diversity challenge in robotics datasets. Using VLMs for robot data augmentation is appealing for two reasons. By leveraging internet-scale pretraining, VLMs are capable of zero-shot generation of contextually appropriate language labels that are also grounded in the physical scenes depicted in demonstration videos. Second, these models can effectively reason about temporal sequences in videos \citep{achiam2023gpt, bai2025qwen25vltechnicalreport, zhang2024videoinstructiontuningsynthetic, gemini25}, enabling segmentation of demonstrations into meaningful sub-tasks. This capability allows for decomposing existing demonstrations into more granular language-action segments, effectively increasing the diversity of our training data without requiring additional data collection.

\begin{figure*}[t]
    \centering
    \includegraphics[width=0.85\textwidth]{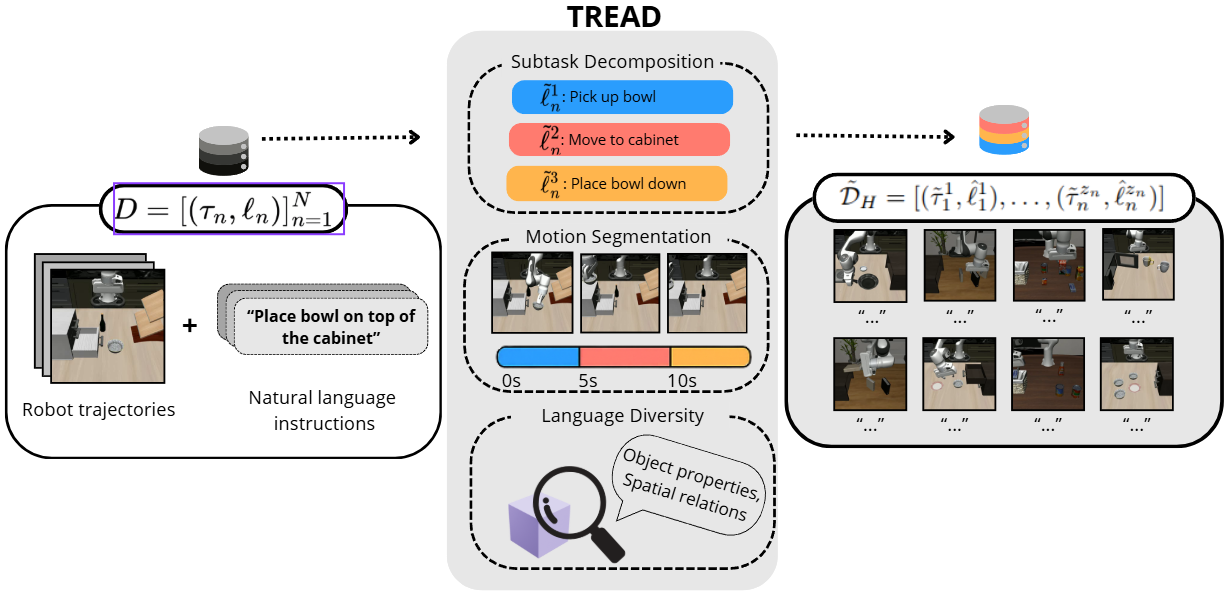}
    \caption{\methodName uses a large-scale VLM to programmatically cut trajectories at sub-goals, label those sub-goals and add variations to the goal-text.}
    \label{fig:method}
\end{figure*}

Motivated by the strong capabilities of large-scale pretrained VLMs, we propose \textbf{T}ask \textbf{R}obustness via R\textbf{E}-Labelling Vision-\textbf{A}ction Robot \textbf{D}ata (\methodName), a framework to augment robotics datasets. \methodName shown in \Cref{fig:method} operates in an iterative three-stage process, where the model’s outputs from previous steps are fed back as inputs to the next: First, a pretrained VLM analyzes the original instruction labels and the initial scenes to generate a sequence of semantic sub-task descriptions that collectively accomplish the original goal. Then the demonstration videos are processed by the the VLM to temporally segment the trajectories, identifying which portions of the demonstration correspond to each sub-task description. Lastly, the VLM generates diverse instructions for these sub-taks incorporating object properties and spatial relationships conditioned on the initial scenes of the sub-tasks and the semantic sub-task descriptions. This temporal-semantic alignment creates multiple language-conditioned sub-demonstrations from each original demonstration from the offline dataset.

In this work, we make two key contributions: (1) a novel framework to utilize iterative querying of a VLM to augment and diversify the original dataset, and (2) the augmented dataset obtained by applying our framework, which will be released to facilitate future research. In our experiments with the LIBERO dataset \cite{liu2023libero} we show that the resulting augmented dataset improves zero-shot performance and text goal following of the vision-language-action policy Octo \cite{octo_2023} and $\pi_0$-FAST \citep{pertsch2025fast}.

%% file: icra_submission/02_related_work.tex
\section{Related Work}

\subsection{Robot learning datasets}
Recently, the robotics community has invested substantial effort into collecting larger manipulation datasets. While earlier datasets such as RT-1 \citep{brohan2022rt}, MT-Opt \citep{kalashnikov2021mt}, and BC-Z \citep{jang2022bc} boast significant numbers of trajectories, they lack diversity in tasks and scenes, which is more important for generalization \citep{khazatsky2024droid, lin2024data, xie2024decomposing}. Responding to this need, several works have focused on increasing dataset diversity by scaling data collection, including BridgeV2 \citep{walke2023bridgedata}, DROID \citep{khazatsky2024droid}, and RDT-1B \cite{liu2024rdt}. These newer datasets not only collect data across more scenes but also demonstrate more complicated tasks consisting of multiple meaningful sub-goals. However, this advancement creates a new challenge: although the number of tasks has increased, the language diversity is diluted by the larger amount of image-action frames per instruction due to longer trajectories. Our work presents a generic framework for expanding the language-image-action distribution of any robotic manipulation dataset without requiring additional data collection, effectively reducing this dilution problem.

\subsection{Vision-language-action models as scalable robot policies}
Vision-Language Models (VLMs) have shown increasing capabilities in complex visual tasks such as state identification, reasoning, and visual question answering \citep{liu2023llava, beyer2024paligemma, achiam2023gpt, gemini25}. Inspired by the success of VLMs, the robotics community adopted similar multi-modal architectures for imitation learning \citep{brohan2022rt, octo_2023, Jiang2023VIMARM, liu2024rdt}. More recent methods start from a pre-trained VLM backbone to leverage internet-scale pretraining \citep{kim2024openvla, black2410pi0}. Followup works have explored architectural modifications \citep{kim2025fine, liu2024rdt, pertsch2025fast} to improve performance in instruction generalization and following. In contrast, our work improves existing policy performance through data augmentation alone. Therefore, our method can be used alongside any policy architecture.

\subsection{Data augmentation methods.}
Data augmentation is a powerful technique for enhancing model robustness and generalization by generating additional synthetic training examples from existing data. In robot learning, augmentation methods have developed along three main dimensions: visual augmentation \citep{bharadhwaj2024roboagent, Yu2023ScalingRL, mandi2022cacti, Chen2023GenAugRB}, language augmentation \citep{zhang2024sprint, Lynch2022InteractiveLT, xiao2022robotic}, and trajectory augmentation \citep{zhang2024sprint, chen2025s2i, raj2024learning}. Our work bridges language and trajectory augmentation by using VLMs to both relabel trajectories and identify meaningful sub-task motions within longer demonstrations. DIAL \citep{xiao2022robotic}, which is most similar to our approach in terms of language augmentation, relies on a predetermined set of labels and requires fine-tuning on related robotics data. In contrast, our method leverages zero-shot VLM capabilities. NILS \citep{blank2024scalingshort} is another framework that incorporates multiple large models and heuristics to segment and relabel trajectories, whereas our method (TREAD) relies on a single iteratively queried VLM without heuristics, making it simpler to reproduce. Lastly, SPRINT \citep{zhang2024sprint}, another closely related work, focuses on composing shorter skills into longer sequences. However, our approach does the opposite by decomposing longer-horizon demonstrations into meaningful sub-tasks. This makes our method particularly suitable for enhancing language-action diversity in open-source robotics datasets.

%% file: icra_submission/03_method.tex
\section{Method}

Here we describe how we design \methodName to segment and stitch robotics data, in three stages: (1) \textit{segmentation} using a VLM to process trajectories for semantic and motion-based properties to find key sub-goal transitions, (2) \textit{labeling} of the new sub-trajectories with sub-goals and to augment the original dataset with language diversity, and (3) \textit{train} a vision-language-action model on the resulting dataset to improve robustness. The pseudocode for \methodName is shown in \Cref{alg:overview}.

\subsection{Problem Formulation}

We assume access to an offline dataset $\mathcal{D}$ of robot trajectories labelled with natural language task instructions. Formally, we define a dataset $\mathcal{D}=[(\tau_n, \ell_n)]^N_{n=1}$ consisting of $N$ labeled trajectories, where each trajectory $\tau_n=[(\mathbf{o}^n_t,\mathbf{a}^n_t)]^T_{t=1}$ contains observation-action pairs over $T$ timesteps. Here, $\mathbf{o}^n_t$ and $\mathbf{a}^n_t$ denote the image observation and action at time $t$ respectively, and $\ell_n$ denotes the natural language instruction describing the task demonstrated in trajectory $\tau_n$. We then train a policy $\pi(\cdotp\mid\mathbf{o}_t,\ell)$ via imitation learning to generate action distributions that mimic demonstrated behavior. Specifically, we look at models trained by minimizing the bahvior cloning objective:
$$ \hat{\pi}^* = \arg\min_\pi \sum_{(\tau,\ell)\in\mathcal{D}}\sum_{(\mathbf{o}_t,\mathbf{a}_t)\in\tau}\mathcal{L}_{BC}(\pi(\cdotp\mid\mathbf{o}_t,\ell),\mathbf{a}_t) $$
where $\mathcal{L}_{BC}$ denotes a supervised loss (e.g., negative log-likelihood) between the predicted action distribution and the demonstrated action.

We are interested in using data augmentation to improve the learned policy's generalization performance. Specifically, we address the question: how can we augment $\mathcal{D}$ with additional trajectory-instruction pairs $(\tau, \ell)$ derived from existing demonstrations, without requiring any new data collection?
\begin{figure*}[h!]
    \centering
    \includegraphics[width=0.7\textwidth]{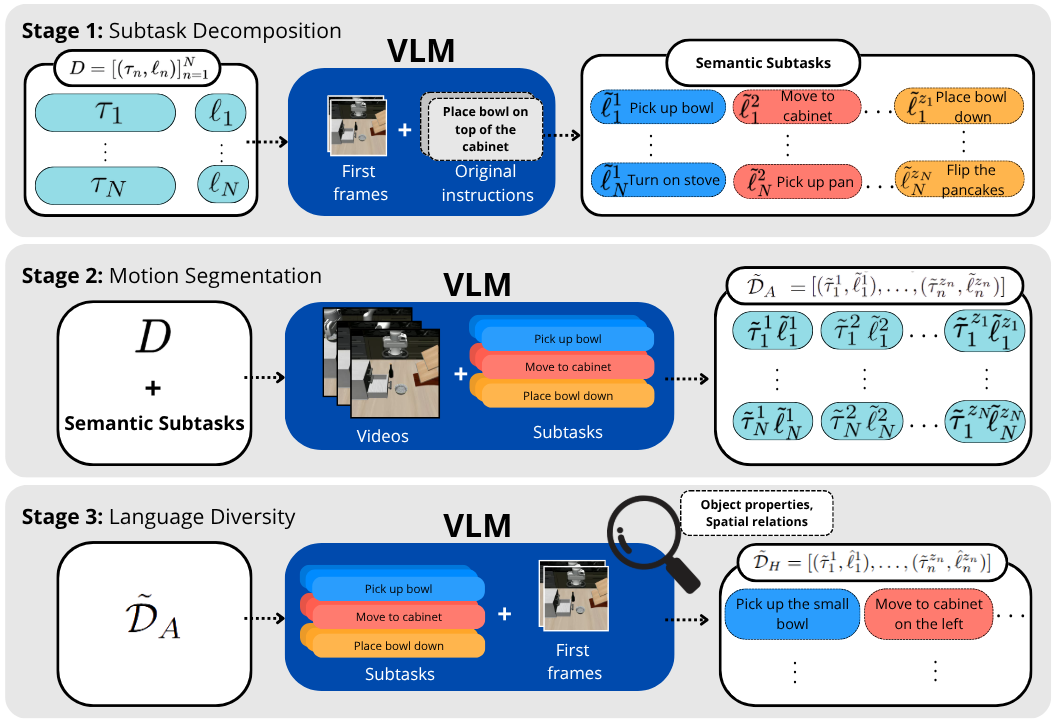}
    \caption{\textbf{TREAD pipeline overview.} Given dataset $\mathcal{D}$ of $N$ labeled trajectories $[(\tau_n, \ell_n)]^N_{n=1}$, \methodName decomposes the dataset into semantically meaningful sub-trajectories through three stages: (1) \textbf{Subtask Decomposition}: Given the original task instruction $\ell$ and initial frame $\mathbf{o}_1$, we prompt the VLM to generate a sequence of sub-task labels $[\tilde\ell^1, \tilde\ell^2,\dots,\tilde\ell^{z_n}]$. (2) \textbf{Motion Segmentation}: The VLM identifies temporal boundaries for each sub-task, resulting in a set of labeled sub-trajectories $[(\tilde\tau^1,\tilde\ell^1),\dots,(\tilde\tau^{z_n},\tilde\ell^{z_n})]$. (3) \textbf{Language Diversity}: The VLM generates diverse paraphrases for each sub-task instruction by leveraging visual context to incorporate object properties (color, shape, material) and spatial relationships (relative positions, orientations).}
    \label{fig:decomp}
\end{figure*}

\subsection{Decomposition of Trajectories}
\label{sec:decompose}

Although our framework does not depend on a specific VLM, in this work we use Gemini Pro 2.5 \citep{gemini25}. Gemini is a multi-modal model that can take as input images, videos, and text, and output text. Given a trajectory $\tau_n$ completing a long-horizon task $\ell_n$, we first derive semantic sub-tasks and then identify which parts of the trajectory complete those sub-tasks. We found that just asking the VLM to segment the video in one-shot provided poor results; providing the model with subtasks resulted in qualitatively better video segmentation. We accomplish this by chaining two VLM inference calls shown in \Cref{fig:decomp}. First, we prompt the VLM to create a plan of sub-task motion labels $[\tilde\ell^1_n, \tilde\ell^2_n, \dots, \tilde\ell^{z_n}_n ]$, given the higher-level task $\ell_n$, and the first frame of the trajectory $\mathbf{o}^n_1$ for grounding and context. Here $z_n$ denotes the number of sub-tasks that are required to complete task $\ell_n$ in the scene as determined by the VLM. For example, the task \textit{``place bowl in the top drawer''} will require a different number of sub-task motions depending on whether the top drawer is currently closed or open and the algorithm should identify this and produce three sub-tasks instead of two. \newline
\vspace{0pt}
Subsequently, we query the model with the video of the trajectory and prompt it to retrieve the starting and ending seconds at which the previously identified sub-tasks occur. As a result, we now have a process by which we can decompose a labeled trajectory $(\tau, \ell)$ into an ordered set of labeled sub-trajectories $[(\tilde\tau^1, \tilde\ell^1), (\tilde\tau^2, \tilde\ell^2), ..., (\tilde\tau^z, \tilde\ell^z)]$,
where $\tilde\tau \subset \tau$ denotes a sub-trajectory. Finally, we can now process $\mathcal{D}$ to construct a new \textit{decomposed} dataset:
$$\tilde{\mathcal{D}}_A=[(\tilde\tau^1_1,\tilde\ell^1_1), \dots, (\tilde\tau_1^{z_1},\tilde\ell_1^{z_1}), \dots, (\tilde\tau^1_n,\tilde\ell^1_n), \dots,  (\tilde\tau_n^{z_n},\tilde\ell_n^{z_n})]$$

For details on the full prompts, see Appendix~\ref{app:prompts}. Practically, we decompose a dataset of 82 unique scene-instruction tasks into 146 subtasks (see \Cref{sec:libero} for details on the dataset).

\subsection{Instruction Diversity and Data Augmentation}

Recent research has highlighted instruction following as a weakness of current VLA models \cite{kim2025fine, pertsch2025fast, liu2024rdt}, attributing the difficulty to the models paying more attention to visual inputs than the language goal. While the mentioned works address the problem algorithmically, we argue that additional textual diversity will help robustness to language. Many robotics datasets have considerably small linguistic diversity which potentially leads to policies being overly sensitive to instruction wording. 
We use the generated sub-trajectories to augment $\mathcal{D}$. 
We consider the following textual augmentation method for language diversity:

 \textbf{Grounded textual diversity.\label{sec:diversify}} We can enhance linguistic diversity by leveraging VLMs to generate visually grounded paraphrases of instructions. For each sub-trajectory $(\tilde\tau_i,\tilde\ell_i)$, we prompt the VLM with both the original instruction $\tilde\ell_i$ and the first visual frame from $\tilde\tau_i$, requesting $k$ language alternatives that preserve the task semantics. As shown in \Cref{fig:decomp}, including the image in the context allows the model to incorporate spatial relationships and perceivable characteristics for more accurate goal diversity beyond just using synonyms. 
 For instance, given an instruction \textit{``pick up the blue coffee mug,"} our VLM might generate alternatives like \textit{``grasp the small coffee mug"} (object attributes) or \textit{``retrieve the coffee mug next to the laptop,"} (spatial relationships). We apply this language augmentation to our already decomposed dataset $\tilde\mathcal{D}_A$ and denote the resulting dataset $\tilde\mathcal{D}_H$:
  $$\tilde{\mathcal{D}}_H=[(\tilde\tau^1_1,\hat\ell^1_1),\dots, (\tilde\tau^{z_1}_1,\hat\ell^{z_1}_1),\dots,(\tilde\tau_n^{1}, \hat\ell_n^{1}),\dots(\tilde\tau_n^{z_n}, \hat\ell_n^{z_n})]$$
 See Appendix~\ref{app:prompts} for the templated prompt we use for grounded textual diversity.

\begin{algorithm}
  \small
  \caption{The \methodName algorithm
    \label{alg:overview}}
   \begin{algorithmic}[1]
    \Require{Robotic dataset $\mathcal{D}$ and VLM $\mathcal{G}$}
    \vspace{5pt}
    \Function{TREAD}{$\mathcal{D}, \mathcal{G}$}
      \Let{$\tilde\mathcal{D}_A$}{\Call{Decompose}{$\mathcal{D}, \mathcal{G}$}}
      \Let{$\tilde\mathcal{D}_H$}{\Call{Diversify}{$\tilde\mathcal{D}_A, \mathcal{G}$}}
      
      \State Train policy $\pi_A$ on mixture of $\mathcal{D}$ and $\mathcal{D}_A$ 
      \State Train policy $\pi_H$ on mixture of $\mathcal{D}$ and $\mathcal{D}_H$
    \EndFunction
    \vspace{2pt}
    \Function{Decompose}{$\mathcal{D}, \mathcal{G}$} \Comment{\Cref{sec:decompose}}
      \Let{$\tilde\mathcal{D}_A$}{\{\}}
      \For{trajectory-instruction pair $(\tau_n,\ell_n) \textrm{ in } \mathcal{D}$}
        \Let{$\mathbf{o}_1$}{first observation in $\tau_n$} \vspace{2pt}
        \Let{$\mathbf{v}$}{$[\mathbf{o}_1, \dots, \mathbf{o}_T]$ (video of trajectory $\tau_n$)} \vspace{2pt}
        \Let{$[\tilde\ell_n^1, \dots, \tilde\ell_n^z]$}{$\mathcal{G}(\mathbf{o}_1,\ell_n)$} \vspace{2pt}
        \Let{$[(\tilde\tau_n^1, \tilde\ell_n^1), \dots, (\tilde\tau_n^z, \tilde\ell_n^z)]$}{$\mathcal{G}(\mathbf{v},[\tilde\ell_n^1, \dots, \tilde\ell_n^z])$} \vspace{2pt}
        \Let{$\tilde\mathcal{D}_A$}{$\tilde\mathcal{D}_A \cup [(\tilde\tau_n^1, \tilde\ell_n^1), \dots, (\tilde\tau_n^z, \tilde\ell_n^z)]$}
      \EndFor
      \State \Return{$\tilde\mathcal{D}_A$}
    \EndFunction
    \vspace{2pt}
    \Function{Diversify}{$\tilde\mathcal{D}_A, \mathcal{G}$} \Comment{\Cref{sec:diversify}}
      \Let{$\tilde\mathcal{D}_H$}{\{\}}
      \For{sub-trajectory--instruction pair $(\tilde\tau^i_n,\tilde\ell^i_n) \textrm{ in } \tilde\mathcal{D}_A$}
        \Let{$\mathbf{o}_1$}{first observation in $\tilde\tau^i_n$}
        \Let{$\hat\ell^i_n$}{choose one from $\mathcal{G}(\mathbf{o}_1, \tilde\ell^i_n)$}
        \Let{$\tilde\mathcal{D}_H$}{$\tilde\mathcal{D}_H \cup (\tilde\tau^i_n, \hat\ell^i_n)$}
      \EndFor
      \State \Return{$\tilde\mathcal{D}_H$}
    \EndFunction
    \vspace{2pt}
  \end{algorithmic}
\end{algorithm}

\subsection{Training a Vision-Language-Action Model}
We study the quality of our dataset augmentation by training vision-language-action manipulation policies on the augmented data. Using the datasets generated by \methodName, we finetune two generalist robot policies: Octo (specifically Octo-Small 1.5) \citep{octo_2023} and $\pi_0$-FAST \citep{pertsch2025fast}. Octo is a diffusion-based policy trained on the Open-X dataset \citep{o2024open}, while $\pi_0$-FAST leverages a pre-trained VLM backbone, PaliGemma \citep{beyer2024paligemma}, trained on the cross-embodied robot data mixture from $\pi_0$ \citep{black2410pi0}, with actions discretized using the FAST tokenization scheme.

To determine the dataset weighting ratio such that the finetuned model achieves the best performance across the original full trajectories ($\mathcal{D}$) and relabeled sub-trajectories ($\tilde\mathcal{D}_A$/$\tilde\mathcal{D}_H$), we conducted ablation studies across multiple mixing ratios. We evaluated ratios ($\mathcal{D}$:$\tilde\mathcal{D}_A$/$\tilde\mathcal{D}_H$) of \textit{1:2}, \textit{1:1.5} and \textit{1:1.1}. The final ratio was selected using Re-Mix \citep{hejna2024reshort}. For all experiments reported in \Cref{sec:exp_setup}, we use the best-performing weighting ratio for each respective data mixture. Detailed performance results for each weighting ratio across the evaluation tasks are provided in Appendix~\ref{app:ratios}.

For all data mixtures, we employ consistent finetuning parameters across models. We finetune Octo for 50,000 steps, batch size of 256,
using a linear warmup followed by cosine decay \citep{loshchilov2017sgdr} with a peak learning rate of $3\times10^{-4}$. For $\pi_0$-FAST, we perform full finetuning for 30,000 steps with a batch size of 32, employing the same cosine decay schedule with warmup but with a peak learning rate of $2.5\times10^{-5}$. Based on test performance, we use checkpoints at step 30,000 for Octo and step 15,000 for $\pi_0$-FAST across all dataset ablations (see \Cref{sec:results}).

\begin{figure*}[h!]
    \centering
    \includegraphics[width=\textwidth]{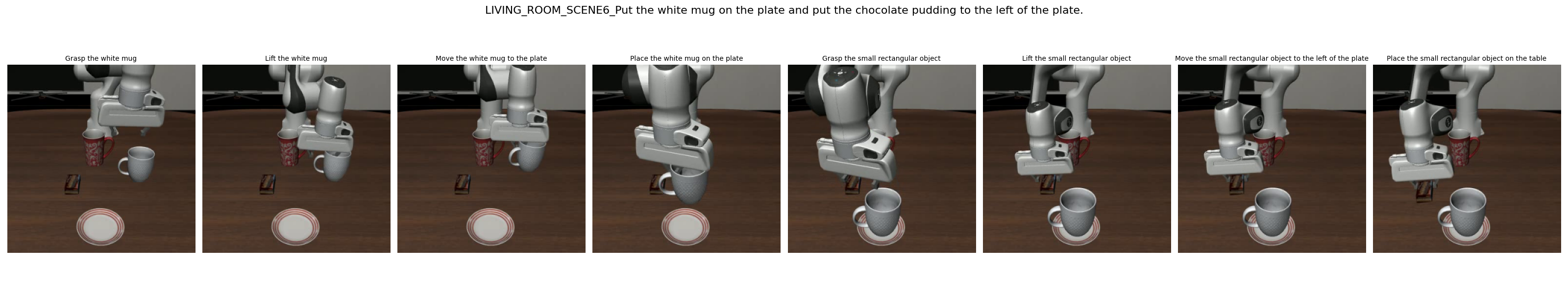}
    \cprotect\caption{Example visualization of the sub-task keypoints produced by \methodName for the trajectory ``\textit{Put the white mug on the plate and put the chocolate pudding to the left of the plate}". Each frame (from left to right) corresponds to the first frame of the sub-task (labeled above). The sub-task labels in order are: (1) ``\textit{Grasp the white mug}" (2) ``\textit{Lift the white mug}" (3) ``\textit{Move the white mug to the plate}" (4) ``\textit{Place the white mug on the plate}" (5) ``\textit{Grasp the small rectangular object}" (6) ``\textit{Lift the small rectangular object}" (7) ``\textit{Move the small rectangular object to the left of the plate}" (8) ``\textit{Place the small rectangular object on the table.}"}
    \label{fig:film-strip1}
\end{figure*}

%% file: icra_submission/04_results.tex
\section{Results}
\label{sec:results}
In our evaluation, we design experiments to investigate the effectiveness of training an agent on a dataset augmented by \methodName toward generalizing to unseen tasks. Specifically, our experiments answer the following research questions: (1) Does \methodName's \textit{trajectory} decomposition increase planning generalization? (2) Does \textit{language} augmentation help policy generalization to text-based goals? 

We focus on robot manipulation and evaluate our method on the image-based dataset LIBERO \citep{liu2023libero}, a simulation benchmark designed for studying lifelong learning in robotic manipulation. The use of LIBERO to evaluate VLA models pretrained on real-world data is a common practice to show the capabilities of these models \cite{pertsch2025fast, kim2024openvla,kim2025fine}. Due to resource constraints, we evaluate our method on a subset of LIBERO as described below.

\subsection{Experimental Setup\label{sec:exp_setup}}
In our experiments, we evaluate the effects of both trajectory decomposition and instruction diversity on policy performance through ablation studies using the following data mixtures:
\begin{enumerate}
    \item \textbf{Finetuned Octo/$\pi_0$-FAST with no augmentations (Original Fine-tuned)}: We finetune the Octo/$\pi_0$-FAST on the unaltered subset of LIBERO-100 trajectories $\mathcal{D}$. This serves as our primary comparison point, expected to perform adequately on in-distribution task-instruction pairs but struggle with language-following in novel scenes or instruction text variations.
    \item \textbf{Finetuned Octo/$\pi_0$-FAST with Trajectory Decomposition (TREAD w/o diverse labels)}: We finetune Octo/$\pi_0$-FAST on a mixture of the unaltered dataset $\mathcal{D}$ and its decomposition $\tilde\mathcal{D}_A$. This ablation isolates the effect of trajectory decomposition, which should help the policy perform tasks in new environments by leveraging subtask compositions seen during training, even when the complete task-environment pairing is novel.
    \item \textbf{Finetuned Octo/$\pi_0$-FAST with Trajectory Decomposition + Label Diversity (TREAD)}: We finetune Octo/$\pi_0$-FAST on a mixture of the unaltered dataset $\mathcal{D}$ and its linguistically enriched decomposition $\tilde\mathcal{D}_H$. This full implementation of our method is expected to handle both novel environments and textual variations by bringing diverse text instructions in distribution.
\end{enumerate}

\textbf{LIBERO. \label{sec:libero}} The LIBERO benchmark \citep{liu2023libero} consists of four task suites that are designed to examine distribution shifts in the object types, the spatial arrangement of objects, the task goals, or the mixture of the previous three. For our experiments, we focus on the LIBERO-100 task suite, which contains 100 tasks involving diverse object interactions and versatile skills with 50 human-teleoperated demonstrations each. In LIBERO-100, tasks are defined by the combination of scene and instruction, meaning that identical instructions performed in different scenes are treated as distinct tasks. Importantly, we utilize the full LIBERO-100 suite rather than the smaller subsets employed by recent works \citep{kim2024openvla,pertsch2025fast}. This larger benchmark presents a more challenging evaluation setting that reduces the risk of overfitting and provides a more rigorous test of our method's ability to improve motion and language generalization (see \Cref{sec:tasks}). \newline
\vspace{0pt}
Due to resource constraints, we labeled five demonstrations per task and omitted any tasks performed in \verb|STUDY_SCENE|s for a total of 570 trajectories. An example of a labeled trajectory is shown in \Cref{fig:film-strip1}.  To evaluate our research questions, we defined two sets of novel tasks within LIBERO:

\subsection{Evaluation Tasks\label{sec:tasks}}
\textbf{Motion Generalization (MG)}: For research question (1), we created new instruction-scene pairs by taking existing language instructions and pairing them with \textit{compatible} but previously unpaired scenes. For example, the instruction \textit{``open the top drawer of the cabinet"} is paired with scene \verb|KITCHEN_SCENE4| containing a cabinet, with this specific combination not appearing in the training data. These tasks will evaluate whether trajectory decomposition helps the agent transfer learned skills to novel environmental contexts. Importantly, these new instruction-scene pairs also evaluate the model's language-following capability, as the model must avoid defaulting to previously trained tasks from those scenes and instead execute the specified instruction. We create 7 such novel scene-instruction environments within the LIBERO framework.
\begin{table*}[ht!]
    \centering
    \small
    \caption{\textbf{LIBERO task performance results.} Success rates (SR) with standard error across the custom evaluation task suites (see \Cref{sec:tasks}) and LIBERO-10 within the LIBERO framework \citep{liu2023libero}. Results include policies fine-tuned on different data mixtures: original trajectory data (Original Fine-tuned), mixture of the unaltered dataset and its decomposition (TREAD w/o div.), and mixture of the unaltered dataset and its linguistically enriched decomposition (TREAD). Bold and underlined values indicate best and second-best performance.}
    \label{tab:results}
    \resizebox{0.78\textwidth}{!}{\begin{tabular}{llccc|ccc}
        \toprule
        \multirow{2}{*}{\textbf{Test Case}} & \multirow{2}{*}{\textbf{Metric (\%)}} 
        & \multicolumn{3}{c|}{$\pi_0$-FAST \citep{pertsch2025fast}} 
        & \multicolumn{3}{c}{Octo \citep{octo_2023}} \\
        \cmidrule(lr){3-5} \cmidrule(lr){6-8}
        & & \shortstack{Original\\ Fine-tuned} & \shortstack{\methodName\\ w/o div.} & \methodName & \shortstack{Original\\ Fine-tuned} & \shortstack{\methodName\\ w/o div.} & \methodName \\
        \midrule

        \multirow{3}{*}{\shortstack{\textbf{Language}\\\textbf{Generalization}}} 
        & Single Goal SR & 47$\pm20$ & \textbf{77}$\pm7$ & \underline{67}$\pm14$ & \underline{82}$\pm2$ & 76$\pm17$ & \textbf{91}$\pm6$ \\
        & 1 of 2 SR       & \underline{63}$\pm11$ & 62$\pm10$ & \textbf{67}$\pm9$ & \underline{76}$\pm6$ & 70$\pm6$ & \textbf{77}$\pm4$ \\
        & 2 of 2 SR       & \underline{36}$\pm8$ & 35$\pm7$ & \textbf{39}$\pm10$ & \underline{30}$\pm7$ & 28$\pm5$ & \textbf{31}$\pm5$ \\
        \midrule

        \multirow{3}{*}{\shortstack{\textbf{Motion}\\\textbf{Generalization}}}   
        & Single Goal SR & 28$\pm18$ & \underline{31}$\pm20$ & \textbf{34}$\pm16$ & 7$\pm4$ & \textbf{27}$\pm10$ & \underline{22}$\pm15$ \\
        & 1 of 2 SR      & 73$\pm3$ & \textbf{82}$\pm11$ & \textbf{82}$\pm9$ & 13$\pm4$ & \textbf{50}$\pm10$ & \underline{43}$\pm3$ \\
        & 2 of 2 SR      & \textbf{7}$\pm3$ & 0$\pm0$  & 0$\pm0$ & 0$\pm0$ & 0$\pm0$ & 0$\pm$0 \\
        \midrule

        \multirow{2}{*}{\textbf{LIBERO-10}}               
        & 1 of 2 SR      & \textbf{83}$\pm9$ & 72$\pm12$ & \underline{74}$\pm8$ & \textbf{76}$\pm5$ & 69$\pm6$ & \underline{72}$\pm4$ \\
        & 2 of 2 SR      & \textbf{57}$\pm10$ & 43$\pm12$ & \textbf{57}$\pm10$ & \underline{40}$\pm7$ & \textbf{41}$\pm4$ & 38$\pm4$ \\
        \midrule
        
        \textbf{Average} & SR & 49$\pm10$ & 50$\pm11$ & 53$\pm10$ & 41$\pm12$ & 45$\pm10$ & 47$\pm11$ \\
        \bottomrule
    \end{tabular}}
\end{table*}

\textbf{Language Generalization (LG)}: For research question (2), we created linguistically modified versions of existing LIBERO-100 tasks. These modifications included variations such as reordering clauses around conjunctions (changing \textit{``turn on the stove and put the moka pot on it"} to \textit{``put the moka pot on the stove and turn it on"}) 
or removing unnecessary object specifiers when context makes them redundant (changing \textit{``close the bottom drawer of the cabinet"} to \textit{``close the drawer"}). These variations assess whether our language augmentation strategies improve policy robustness to variations in instruction wording. We found 14 existing tasks within LIBERO-100 that satisfied our criteria for instruction modification.

We evaluate all dataset ablations across both test cases. For each task, we conduct $30$ policy rollouts and report the average success rate. To provide a more nuanced understanding of policy capabilities, we separately analyze performance on single-goal tasks (e.g. \textit{``close the drawer"}) and two-goal tasks (e.g. \textit{``put the bowl in the drawer of the cabinet and close it"}). For two-goal tasks, we report both partial success (completing only one goal) and complete success (achieving both goals).

\subsection{Increased Zero-Shot Performance}

The Motion Generalization results in \Cref{tab:results} demonstrate \methodName's effectiveness when policies encounter familiar instructions in novel environments. Both \methodName and \methodName w/o diverse labels outperform Original Fine-tuned across Single Goal tasks, with \methodName achieving 15\% and 6\% higher
success rates on Octo and $\pi_0$-FAST respectively. This shows that the automatic segmentation of \methodName is able to provide more trajectory and goal diversity which drives additional generalization across more tasks.  

The benefits of trajectory decomposition become more pronounced for complex, multi-step tasks. In two-goal scenarios, \methodName achieves substantially higher first-goal completion rates - 30\% improvement for Octo and 9\% for $\pi_0$-FAST - compared to baseline fine-tuning. The wider performance gap for multi-goal tasks likely stems from the increased opportunities for sub-trajectory overlap between training and test scenarios. Notably, the performance gains are more substantial for Octo, which lacks the pretrained vision-language understanding of $\pi_0$-FAST, suggesting that trajectory decomposition provides particularly valuable inductive biases for models without built-in multimodal reasoning capabilities. 


\subsection{Better Text Goal Following}

The Language Generalization results in \Cref{tab:results} reveal \methodName's ability to handle linguistic variations of familliar task-instructions in known environments. In single-goal tasks, \methodName demonstrates substantial improvements over Original Fine-tuned baselines, achieving 20\% higher success rates with Octo and 9\% with $\pi_0$-FAST. 
This improvement provides strong evidence that our grounded textual diversity approach effectively prepares policies for the natural linguistic variations. 

Interestingly, the performance benefits for Language Generalization are most apparent in shorter, single-goal tasks rather than complex multi-step scenarios. This pattern likely reflects the mismatch between the relatively concise language instructions in our decomposed sub-trajectories and the longer, more complex instructions typical of multi-goal tasks. Additionally, longer-horizon tasks provide more opportunities for compounding errors when policies encounter out-of-distribution language conditioning, making it inherently more challenging to maintain robust language following throughout extended task execution.


Importantly, we verify that our augmentation approach does not compromise performance on in-distribution tasks. When evaluated on the in-distribution long-horizon tasks in LIBERO-10, \methodName maintains comparable performance to Original Fine-tuned for two-goal success rates as seen in the last row in \Cref{tab:results}.

%% file: icra_submission/05_conclusions.tex
\section{Conclusion}

In summary, we presented \methodName, a framework that uses off-the-shelf VLMs to perform a type of trajectory segmentation and then stitching via training a large-scale BC model Octo \citep{octo_2023}. We find that fine-tuning Octo on our new dataset improves the performance, especially on more difficult longer horizon tasks and novel tasks not seen during training. Despite these promising results, our work has limitations that point to interesting directions for future research. Our work uses a large closed-source VLM, which limits reproducibility. An important next step would be to evaluate \methodName with open-source alternatives \citep{liu2023llava, bai2025qwen25vltechnicalreport, zhang2025videollama3frontiermultimodal}
and potentially fine-tune these models specifically on robotics data to improve their embodied understanding. Future work can also explore applying \methodName to larger datasets and larger models. However, our work shows that even with small amounts of properly segmented and augmented data there are obvious performance gains for current VLA models.

%% file: icra_submission/06_appendix.tex
\begin{appendices}

\section{VLM Prompts}
\label{app:prompts}
\begin{figure*}[h!]
    \centering
    \includegraphics[width=0.83\textwidth]{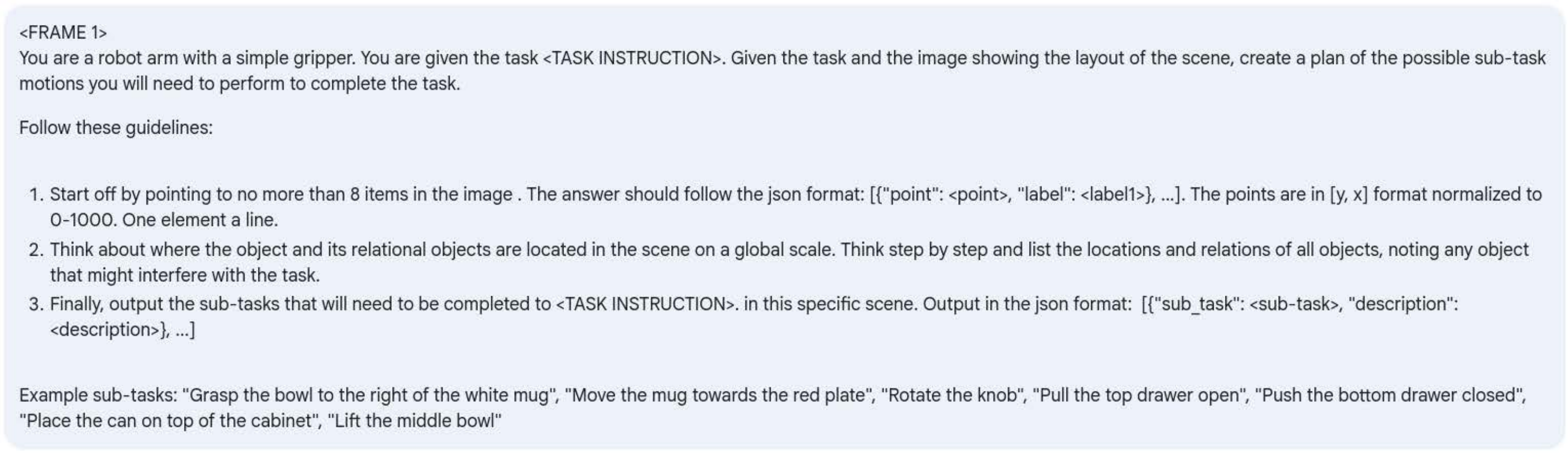}
    \cprotect\caption{The templated prompt we use for generating semantic sub-tasks. The placeholders \verb|<FRAME 1>| and \verb|<TASK INSTRUCTION>| are replaced with the first image of the trajectory and the trajectory instruction, respectively.}
    \label{fig:prompt_1}
\end{figure*}

\begin{figure*}[h!]
    \centering
    \includegraphics[width=0.83\textwidth]{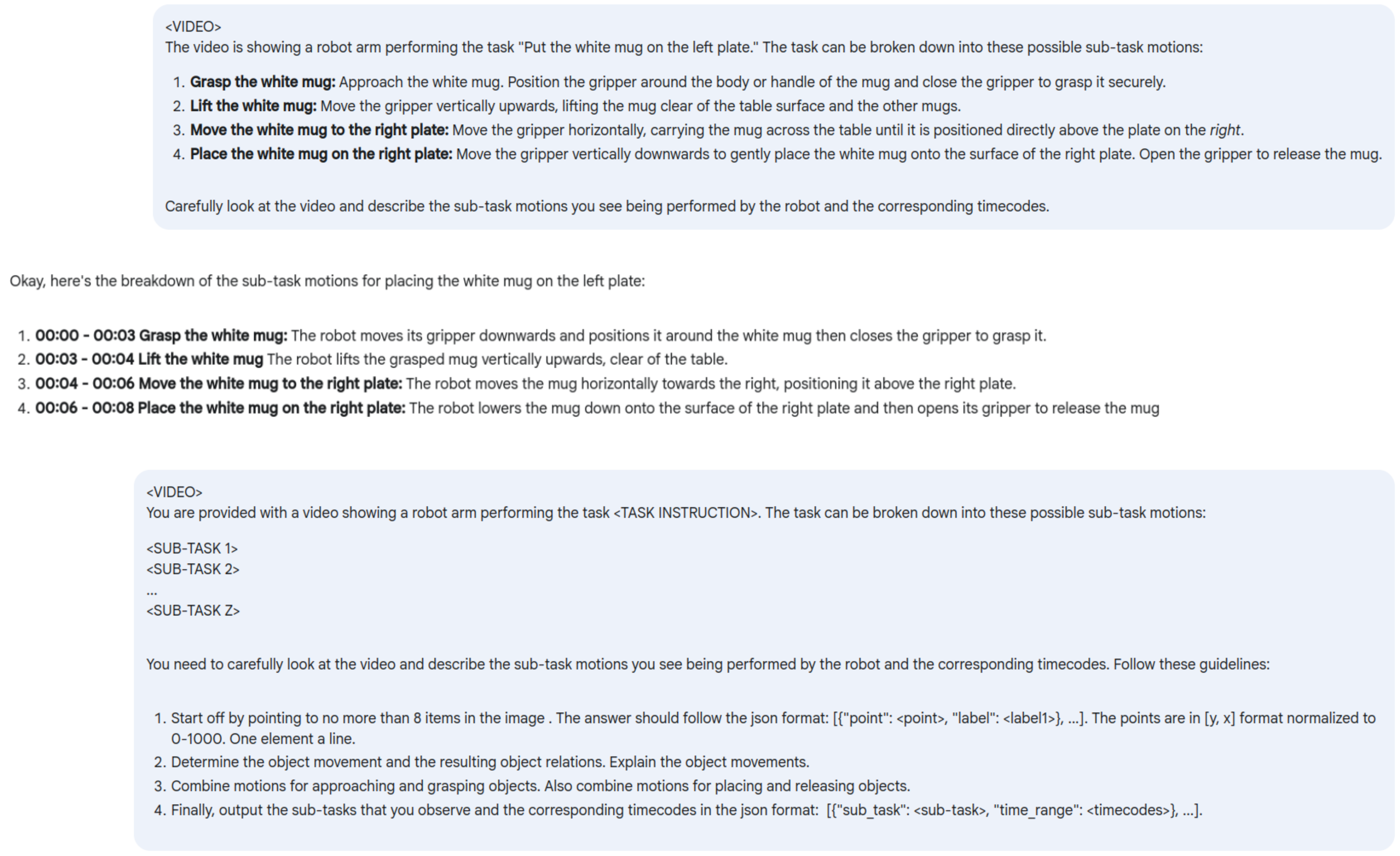}
    \cprotect\caption{The templated prompt we use for recognizing semantic sub-tasks in a trajectory video. We use few-shot prompting by providing the model with 3 example user-model exchanges before the final query prompt (we include only one example for brevity). The placeholders \verb|<VIDEO>| and \verb|<TASK INSTRUCTION>| are replaced with the trajectory video and the trajectory instruction, respectively.}
    \label{fig:prompt_2}
\end{figure*}

\begin{figure*}[h!]
    \centering
    \includegraphics[width=0.83\textwidth]{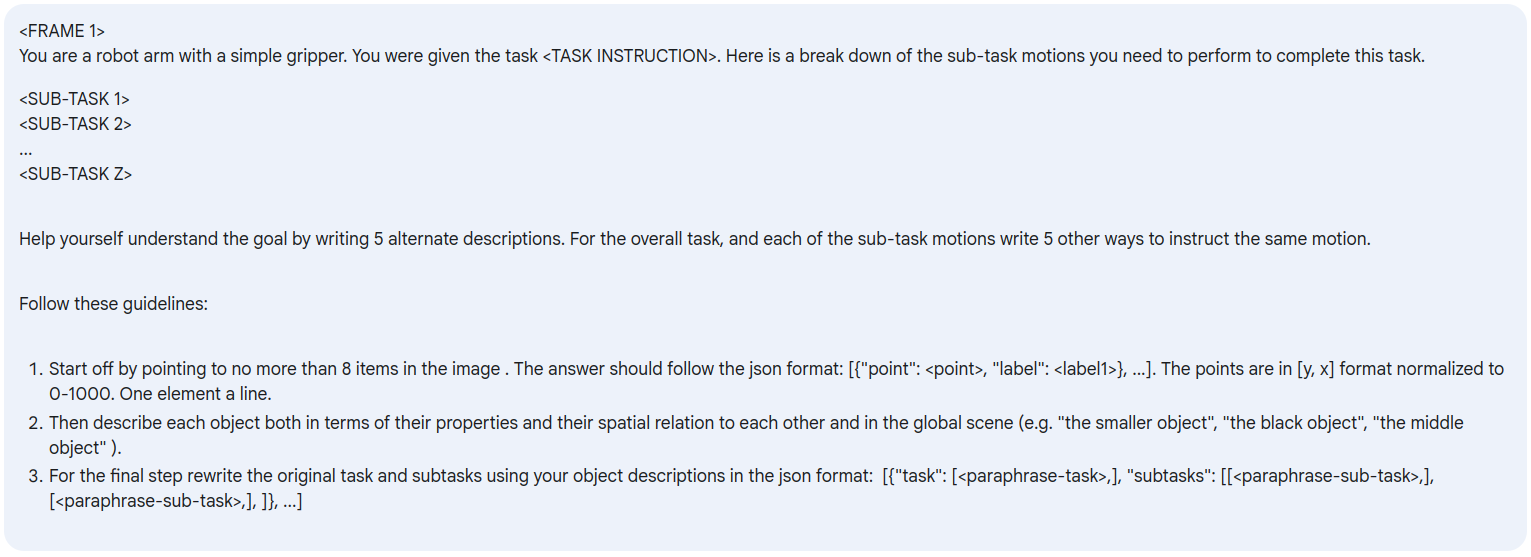}
    \cprotect\caption{The templated prompt we use for grounded textual diversity (see \Cref{sec:diversify}). The placeholders \verb|<FRAME 1>| and \verb|<TASK INSTRUCTION>| are replaced with the first image of the trajectory and the trajectory instruction, respectively.}
    \label{fig:prompt_3}
\end{figure*}

We provide the templated prompts we use for querying Gemini 2.5 \citep{gemini25} for semantic sub-task breakdown and motion identification in  \Cref{fig:prompt_1}, \Cref{fig:prompt_2}, and the template prompt for grounded paraphrasing in \Cref{fig:prompt_3}. The three example tasks used for few-shot prompting motion identification are taken from the LIBERO-90 dataset.

\section{Dataset Weighting Ratios}
\label{app:ratios}
To determine the optimal balance between original demonstrations and augmented sub-trajectories, we conducted ablations across different dataset mixing ratios. \Cref{tab:sample} presents results for Octo, showing success rate performance across three task sets as described in \Cref{sec:tasks}.

\begin{table}[!h] 
    \centering
    \Large
    \caption{Success rate performance of Octo \citep{octo_2023} finetuned with different dataset weighting ratios for (1) TREAD w/o diverse labels ($\mathcal{D}$:$\tilde\mathcal{D}_A$) and (2) TREAD with diverse labels ($\mathcal{D}$:$\tilde\mathcal{D}_H$).}
    \label{tab:sample}

 \resizebox{0.48\textwidth}{!}{  \begin{tabular}{ll|cc|cc|cc}
        \toprule
        \multirow{2}{*}{\textbf{Test Case}} & \multirow{2}{*}{\textbf{Metric (\%)}} 
        & \multicolumn{2}{c|}{\textbf{1:2}} 
        & \multicolumn{2}{c|}{\textbf{1:1.5}} 
        & \multicolumn{2}{c}{\textbf{Re-Mix}} \\
        \cmidrule(lr){3-4} \cmidrule(lr){5-6} \cmidrule(lr){7-8}
        & & \shortstack{TREAD\\ w/o div.} & TREAD 
          & \shortstack{TREAD\\ w/o div.} & TREAD 
          & \shortstack{TREAD\\ w/o div.} & TREAD \\
        \midrule

        \multirow{3}{*}{\shortstack{\textbf{Language}\\\textbf{Generalization}}} 
        & Single Goal SR & 73 & 89 & 51 & 69 & 76 & 73 \\
        & 1 of 2 SR       & 74 & 79 & 72 & 81 & 72 & 81 \\
        & 2 of 2 SR       & 30 & 30 & 32 & 33 & 30 & 36 \\
        \midrule

        \multirow{3}{*}{\shortstack{\textbf{Motion}\\\textbf{Generalization}}}   
        & Single Goal SR  & 22 & 20 & 27 & 23 & 25 & 17 \\
        & 1 of 2 SR       & 42 & 40 & 38 & 40 & 49 & 31 \\
        & 2 of 2 SR       & 0 & 0 & 0 & 0 & 0 & 0 \\
        \midrule

        \multirow{2}{*}{\textbf{LIBERO-10}}               
        & 1 of 2 SR       & 69 & 77 & 68 & 71 & 74 & 70 \\
        & 2 of 2 SR       & 37 & 41 & 37 & 37 & 41 & 37 \\
        \bottomrule
    \end{tabular}}

\end{table}




\end{appendices}